\documentclass[a4paper,twoside]{article}

\usepackage{epsfig}
\usepackage{subcaption}
\usepackage{calc}
\usepackage{amssymb}
\usepackage{amstext}
\usepackage{amsmath}
\usepackage{amsthm}
\usepackage{multicol}
\usepackage{pslatex}
\usepackage{apalike}
\usepackage{algorithm2e}
\usepackage[bottom]{footmisc}
\usepackage{SCITEPRESS}
\usepackage{cite}
\usepackage{amsmath,amssymb,amsfonts}
\usepackage{algorithmic}
\usepackage{graphicx}
\usepackage{textcomp}
\usepackage{xcolor}
\usepackage{epigraph} 
\usepackage{multirow} 
\usepackage{comment}
\usepackage{multirow}
\usepackage{booktabs}
\usepackage{subfig}
\usepackage{float}
\usepackage{stfloats}
\usepackage{flushend}
\usepackage{soul}

\begin{document}

\title{Intrinsic Evaluation of RAG Systems for Deep-Logic Questions\thanks{This work was supported in part by Librum Technologies, Inc.}}

\author{\authorname{Junyi (Edward) Hu\orcidAuthor{0000-0001-8524-0123}, You Zhou\orcidAuthor{0009-0005-0919-5793} and Jie Wang\orcidAuthor{0000-0003-1483-2783}}
\affiliation{Miner School of Computer \& Information Sciences, University of Massachusetts, Lowell, MA, USA}
\email{\{junyi\_hu, you\_zhou1\}@student.uml.edu, jie\_wang@uml.edu}
}


\abstract{We introduce the Overall Performance Index (OPI), an intrinsic metric to evaluate retrieval-augmented generation (RAG) mechanisms for applications involving deep-logic queries. OPI is computed as the harmonic mean of two key metrics: the Logical-Relation Correctness Ratio and the average of BERT embedding similarity scores between ground-truth and generated answers.
We apply OPI to assess the performance of LangChain, a popular RAG tool, using a logical relations classifier fine-tuned from GPT-4o on the RAG-Dataset-12000 from Hugging Face. Our findings show a strong correlation between BERT embedding similarity scores and extrinsic evaluation scores. Among the commonly used retrievers, the cosine similarity retriever using BERT-based embeddings outperforms others, while the Euclidean distance-based retriever exhibits the weakest performance. Furthermore, we demonstrate that combining multiple retrievers, either algorithmically or by merging retrieved sentences, yields superior performance compared to using any single retriever alone.
}

\onecolumn \maketitle \normalsize \setcounter{footnote}{0} \vfill

\section{\uppercase{Introduction}}
\label{sec:introduction}

A RAG system typically consists of two major components: Indexing and Retrieval.
The former is responsible for indexing a reference text document before any queries are made to it.
The latter is responsible for retrieving relevant data from the indexed document in response to a query and passing that information, along with the query, to a large language model (LLM) to generate an answer.
The Retrieval component is typically a framework that supports a variety of retrieval methods, each referred to as a retriever. 

To assess the effectiveness of a retriever in uncovering the logical relationship for an answer to a query with respect to the reference document, we introduce the Overall Performance Index (OPI). This metric measures both the correctness of the answers generated by an LLM and the accuracy of the logical relations produced by a classifier. The OPI is calculated as the harmonic mean of the BERT embedding similarity between ground-truth and generated answers, and the logical-relation correctness ratio.

To demonstrate the effectiveness of the OPI metric, we use the RAG-Dataset-12000 provided by Hugging Face \cite{huggingface_rag_dataset_12000} as the training and testing dataset. We fine-tune GPT-4o to construct a classifier to generate logical relations between a query and an answer, with respect to the reference document.
We then evaluate LangChain \cite{langchain_website}, a popular RAG tool, with seven common retrievers, extracting relevant sentences from the reference document for each query. Using GPT-4o as the underlying LLM, we generate an answer to the query and use the fine-tuned GPT-4o classifier to generate a logical relation. 

To rank retrievers, we calculate the average OPI score across all 13 logical relations provided in RAG-Dataset-12000. 
We then use OPI to analyze the strengths and weaknesses of individual retrievers.
Moreover, we demonstrate that several variations of combining multiple retrievers, either algorithmically or by merging retrieved sentences, outperform a single retriever alone.

\section{\uppercase{{Preliminaries}}}
\label{sec:preliminaries}\label{sec:2}

The technique of RAG was introduced by Lewis et al. (2020) \cite{lewis2020retrieval} 
a few years before the widespread adoption of LLMs. 
The performance of a RAG system relies on the quality of the underlying retriever and the ability of the underlying LLM. 

LangChain is a popular RAG tool, which divides a reference document into overlapping text chunks of equal size. The suffix of each chunk overlaps with the prefix of the next.

To the best of our knowledge, no previous research has comprehensively evaluated the performance of RAG systems in the context of deep-logic question answering.


Given below are seven common sentence retrievers supported by LangChain:

\textbf{DPS} (dot-product similarity) 
converts a query and a text chunk as BERT-based \cite{devlin2018bert} embedding vectors and compute their dot product as a similarity score. It returns $k$ chunks with the highest scores to the query.
(DPS in LangChain is referred to as Cosine Similarity.)

\textbf{kNN} ($k$-Nearest Neighbors) 
in LangChain is the normalized dot-product similarity by the L2-norm, which is widely referred to as the cosine similarity. It returns $k$ chunks with the highest cosine similarity scores to the query.

\textbf{BM25} \cite{robertson2009probabilistic} is a probabilistic information retrieval model that ranks documents based on the term frequency in a chunk and the inverse chunk frequency in the reference document. Let \( q \) be a query, \( T \) a chunk of text, \( f(t_i, T) \) the frequency of term \( t_i \) in \( T \), \( |T| \) the size of $T$, \( \text{avgTL} \) the average chunk length, $N$ the total number of chunks, and $n(t_i)$
        the number of chunks that contain $t_i$. Then BM25$(q,T)$ is defined by
\begin{align*}
        \text{BM25}(q,T) =& \sum_{i=1}^{n} \ln \left(\frac{N-n(t_i) + 0.5}{n(t_i) +0.5}+1\right)\cdot \\
        & \frac{ (f(t_i, T) \cdot (\kappa + 1))}{f(t_i, T) + \kappa \cdot (1 - b + b \cdot \frac{|T|}{\text{avgTL}})},
\end{align*}
where $\kappa$ and $b$ are parameters.        
Return $k$ chunks of text with the highest BM25 scores to the query.

\textbf{SVM} (Support Vector Machine) \cite{cortes1995support} is a supervised learning model that finds the hyperplane that best separates data points in a dataset. To use SVM as a retriever, first represent each chunk of text as a feature vector. This can be done using word embeddings, TF-IDF, or any other vectorization method. Then use the labeled dataset to train an SVM model. Convert the query into the same feature vector space as the chunks. Apply the SVM model to the query vector to produce a score that indicates how similar the query is to each chunk. Extract $k$ chunks with the highest scores.
      

\textbf{TF-IDF} \cite{sammut2011tfidf} measures the importance of a word in a chunk of text relative to the set of chunks in the reference document, combining term frequency and inverse chunk frequency. In particular,
        \[
        \text{TF-IDF}(t,T) = \text{TF}(t,T) \times \text{IDF}(t),
        \]
        where \( t \) is a term, \( T \) is a chunk, and \( \text{IDF}(t) \) is the inverse chunk frequency of \( t \).
        Given a query $q$, select $k$ chunks with the highest TF-IDF$(q,T)$ values. 


\textbf{MMR} \cite{carbonell1998use} 
is a retrieval algorithm that balances relevance and diversity in the selection of $k$ chunks. It iteratively selects chunks that are both relevant to the query and minimally redundant with respect to the chunks already selected.
        
 
{\textbf{EDI} (Euclidean Distance) \cite{bishop2006pattern}
measures the straight-line distance between a query and a chunk, represented in bag-of-words vectors. Return $k$ chunks with the shortest distance to the query. 
\begin{table*}[!tb]
    \centering
    \caption{Information of logical relations} 
    \label{table:LRD}
    \includegraphics[width=\textwidth]{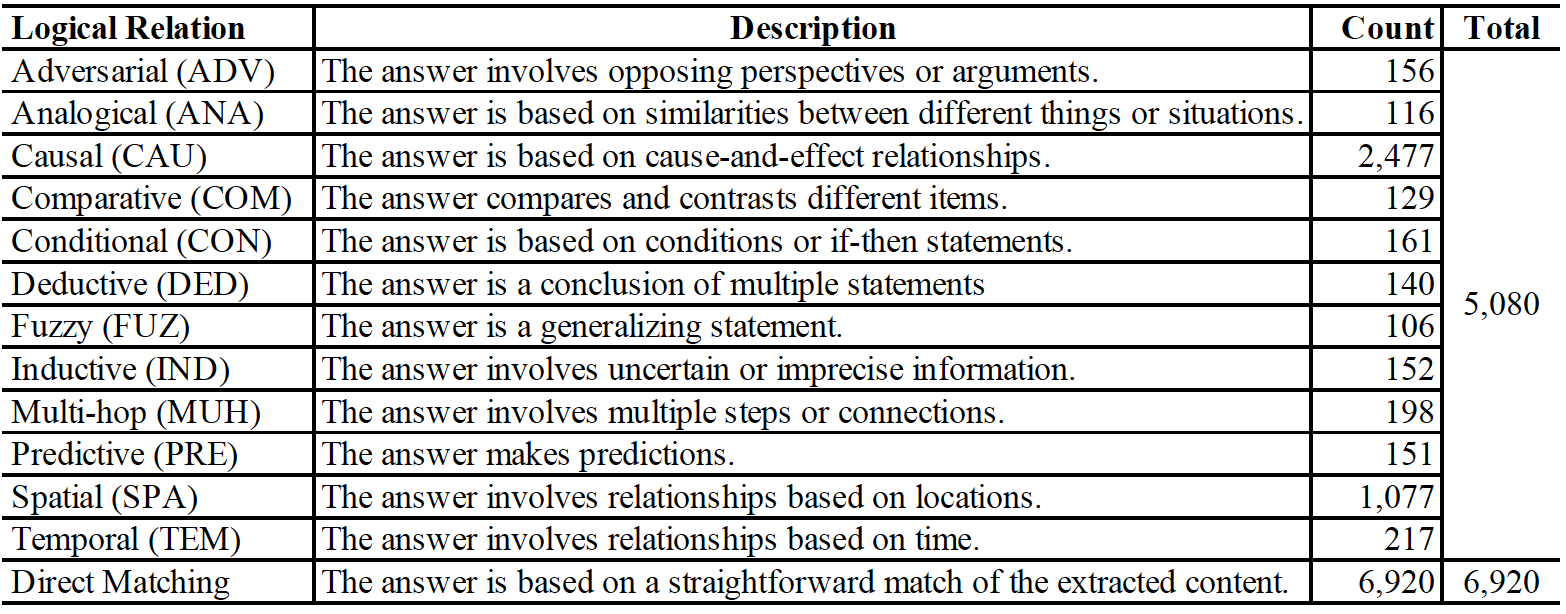}
\end{table*}

A data point in RAG-Database-12000 contains the following
attributes: `context', 
`question', 
`answer', 
`retrieved\_sentences', 
`logical\_relation', 
where `context' is the reference document.
There are thirteen categories of logical reasoning in the dataset. Their names, abbreviations, descriptions, and the distribution of counts are presented in Table \ref{table:LRD}.  All but the last category involve deep logical reasoning, meaning that arriving at the correct answer requires complex, multi-step processes involving multiple concepts, facts, or events extracted from the content. The table includes eleven specific types of deep reasoning, with an additional category for general deep reasoning, referred to as multi-hop reasoning.

\section{\uppercase{Overall Performance Index}}
\label{sec:OPI}

Let $A$ and $LR$ denote, respectively, the ground-truth answer and logical relation to the question with respect to the question $Q$, the context $C$, and the retrieved sentences $S$. Let $A'$ and $LR'$ denote, respectively, the answer and the logical relation generated by a RAG system with an LLM. We represent $A$ and $A'$ using BERT embeddings and compute the cosine similarity of the embeddings.

For a given dataset $D$ with respect to a particular logical relation $LR$, let
BERTSim$_D$ denote the average BERT similarity scores of all $(A,A')$ pairs
and LRCR$_D$ (logical-relation correctness ratio) denote the proportion of data points where the predicted logical relation matches $LR$. Namely,
\begin{align}
    \text{LRCR}_D &= \frac{{|\{d \in D \mid LR = LR'\}|}}{|D|}
\end{align}

The OPI for dataset $D$ is defined by the following parameterized harmonic mean of BERTSim$_D$ and LRCR$_D$, similar to defining the F-measure \cite{lewis1994study}.
\begin{align}
\text{OPI}(\beta)_D &= \frac{(1+\beta^2) \cdot \text{BERTSim}_D\cdot \text{LRCR}_D}{(\beta^2 \cdot \text{BERTSim}_D) + \text{LRCR}_D}.
\end{align}

$\text{OPI}(1)_D$ weighs answer accuracy and logical relation accuracy equally.
$\text{OPI}(\beta)_D$ weighs answer accuracy more heavily  when $\beta > 1$ (e.g., 
$\beta = 2$),
and weighs logical relation accuracy more heavily when $\beta < 1$ (e.g., 
$\beta = 0.5$). 

When there is no confusion in the context, the subscript $D$ is omitted. Denote $\text{OPI}(1)$ as OPI-1, $\text{OPI}(2)$ as OPI-2, and $\text{OPI}(0.5)$ as OPI-0.5.

In addition to BERTSim, other metrics may be used to measure the similarity between the generated answer and the ground-truth answer, such as Hugging Face's MoverScore, as applied in the study of content significance distributions of text blocks in a document \cite{Zhou-Wang2023}. We choose BERTSim because MoverScore uses IDF to compute word weights, which is better suited for extractive answers but less appropriate for generative answers produced by LLMs. 

Experimental results show that the BERTSim metric aligns well with the outcomes of extrinsic comparisons of the ground-truth answers with the generated answers (see Section \ref{sec:extrinsic} for details). 

In what follows, we will use OPI-1 as the default intrinsic measure to study the performance of RAG systems for answering deep-logic questions.


\section{\uppercase{Evaluation}}
\label{sec:evaluation}\label{sec:experiment}




As seen in Table \ref{table:LRD}, the data points in RAG-Dataset-12000 are unevenly distributed across the 13 logical relations, with significant disparities, such as only 106 data points in Fuzzy Reasoning compared to 6,920 data points in Direct Matching. To fine-tune GPT-4o and construct a classifier for identifying logical relations, a balanced dataset is preferred. To achieve this, we randomly select 100 data points from each logical relation category, forming a new dataset called RAG-QA-1300 that consists of 1,300 data points. This dataset is then split with an 80-20 ratio to create a training set and a test set. 

Fine-tuning was performed by combining the context, question, and answer from each data point into a cohesive input text, labeled with its corresponding logical relation. The process involved approximately 800 training steps, resulting in a validation loss of $10^{-4}$. 
This specific checkpoint was selected for its optimal performance.

The fine-tuned GPT-4o classifier for logical relations significantly improves the accuracy to 75.77\% on the test set, compared to 49.23\% when using GPT-4o out-of-the-box without fine-tuning.

We used LangChain with the seven common retrievers mentioned in
Section \ref{sec:2}. We used GPT-4o to generate answers and the fine-tuned GPT-4o classifier to generate logical relations. LangChain supports a wide range of retrievers and allows for the seamless integration of pre-trained LLMs.

\begin{table*}[]
\centering
\caption{Intrinsic comparisons across all logical relations, where ``Retr" is an abbreviation of Retriever, ``B" stands for BERTSim, ``L" for LRCR, and ``O-1" for OPI-1}
\label{table:intrinsic evaluation}
\includegraphics[width=\textwidth]{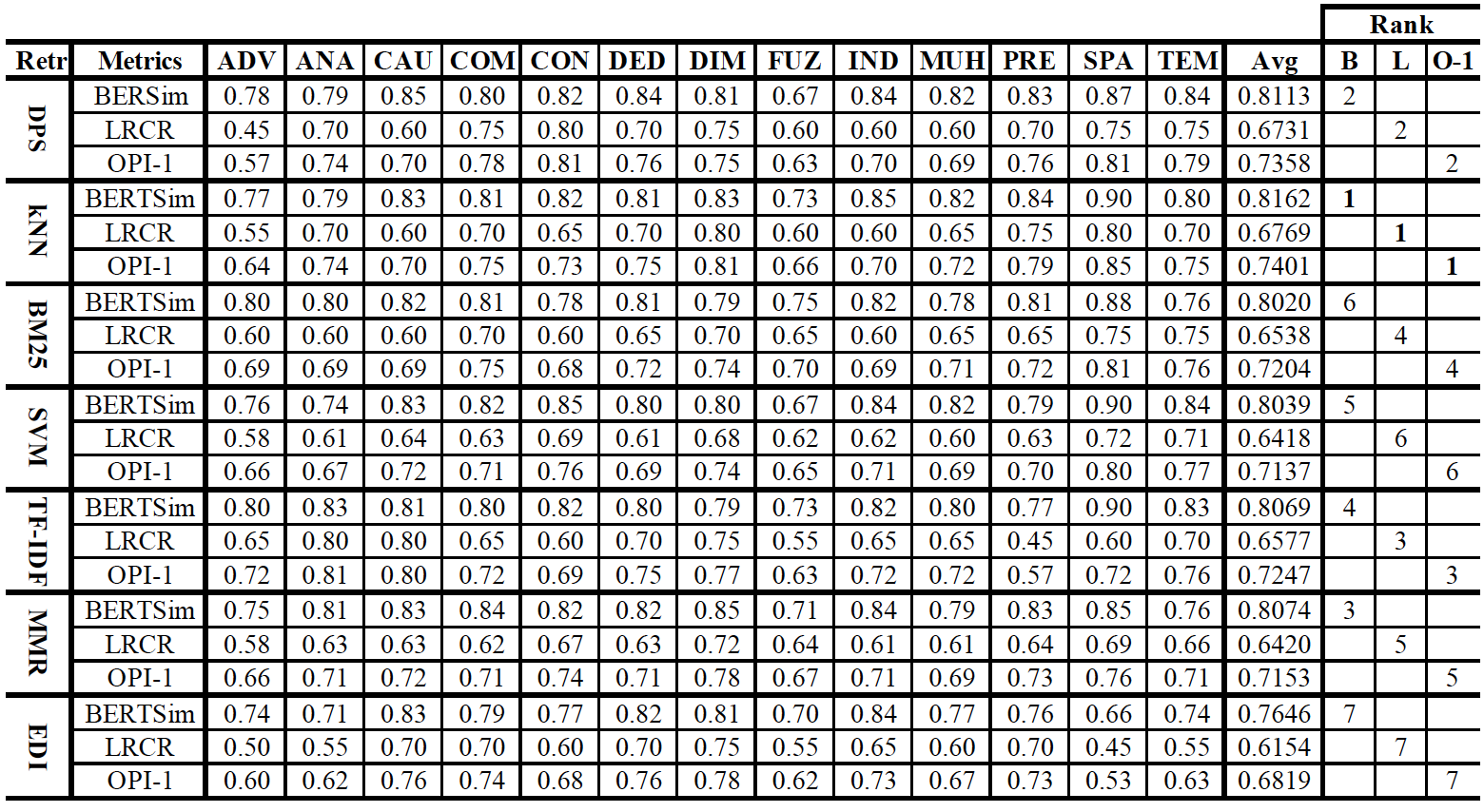}
\end{table*}

\subsection{Intrinsic Evaluation}

We set the chunk size to 100 (words) with a chunk overlap of 20 \% in the setting of LangChain, where paragraph breaks, line breaks, periods, question marks, and exclamation marks are set to be the separators.
These settings were fed into the LangChain function \verb|RecursiveCharacterTextSplitter| to split a reference document into chunks, where each chunk contains up to 100 words, ending at a specified separator to break naturally such that the chunk is as large as possible, and adjacent chunks have a 20\% overlap.

We used the default settings for each retriever to return four chunks in the context with the best scores---highest for similarity and ranking measures, smallest for distance measures---from the underlying retriever as the most relevant to the query. We then converted the four chunks extracted by the retriever back into complete sentences as they appeared in the original article. These sentences and the query were then fed to GPT-4o to generate an answer. Moreover, we instructed GPT-4o to determine the logical relationship for the answer with respect to the input text.

We consider the accuracy of the generated answers and logical relations to be equally important.
Table \ref{table:intrinsic evaluation} 
presents the evaluation results of OPI-1
on the test data of RAG-QA-1300. The OPI-1 score with respect to each retriever is calculated for each set of data points of the same logical relation.
The average OPI-1 score for each retriever across all logical relations is calculated by
%
\begin{align}
\text{OPI-1} &= \frac{2/|L| \cdot \sum_{\ell \in L} \text{BERTSim}_{\ell} \cdot \sum_{\ell\in L}\text{LRCR}_{\ell}}{\sum_{\ell \in L} \text{BERTSim}_{\ell} + \sum_{\ell \in L}\text{LRCR}_{\ell}}, \label{eq:2}
\end{align}
where $L$ is the set of the 13 logical relations, and $\text{BERTSim}_{\ell}$ and $\text{LRCR}_{\ell}$ denote, respectively,
the corresponding BERTSim score and LRCR value for the logical relation 
$\ell$.

An alternative is to calculate the average OPI-1 score across all logical relations. While this differs slightly from Formula (\ref{eq:2}), the difference is minimal. We prefer Formula (\ref{eq:2}) for practical efficiency, as it bypasses the need to compute individual OPI-1 scores for each logical relation when these scores are not needed in applications, streamlining the process and reducing unnecessary computations.

\subsection{Extrinsic Evaluation}
\label{sec:extrinsic}

The extrinsic evaluation uses a 0-3-7, 3-point scoring system to score $A'$ for each pair $(A, A')$, where $A$ is the ground-truth answer and $A'$ is the answer generated directly by GPT-4o based on the question $Q$ and the extracted sentences in the corresponding data point. The scoring system works as follows:
\begin{itemize}
\item $A'$ receives 7 points if it exactly matches the meaning of $A$.
\item $A'$ receives 3 points if it partially matches the meaning of $A$, with or without extra information not contained in $A$.
\item $A'$ receives 0 points if there is no meaningful match.
\end{itemize}
This scoring system was designed to simplify extrinsic evaluation while widening the gaps between perfect, partial, and no matches, providing a more nuanced assessment compared to the traditional 1-2-3, 3-point scoring system. The intention is to create a ranking with greater discrepancy between scores, enabling clearer differentiation of performance levels and allowing for more meaningful comparisons. 

Given below are examples of each scenario.

\begin{itemize}
    \item 7 points: $A'$ receives 7 points if $A$ and $A'$ have exactly the same meaning, with neither additional nor missing information. For example, the following $A'$ using DPS is a 7-point answer.
    \begin{itemize}
        \item $Q$: What is the debate about what Desmond said in the context?
        \item $A$: The debate is about what Desmond said to Locke, specifically whether he said ``snowman" or ``silhouette".
        \item $A'$: The debate is about whether Desmond said ``snowman" or ``silhouette".
    \end{itemize}
    
    \item 3 points. $A'$ receives 3 points in either of the following two cases: 
    
    Case 1: Extra information. $A'$ matches the meaning of $A$, but includes additional information not present in $A$. For example, the following $A'$ using MMR receives 3 points for the reason that $A'$ contains extra information that the Ford GT40 was produced in limited runs from 1964 to 1969.
         \begin{itemize}
                \item $Q$: What was the reason behind the creation of the Ford GT40?
                \item $A$: The Ford GT40 was created as a result of a grudge between Henry Ford II and Enzo Ferrari. After a failed attempt to acquire Ferrari, Henry Ford II decided to compete against Ferrari in the prestigious 24 Hours of Le Mans race. To break Ferrari's dominance in the race, Ford's chief engineer, Roy Lunn, came up with the concept of the GT40.               
                \item $A'$: The Ford GT40 was born out of a rivalry between Henry Ford II and Enzo Ferrari. After a failed attempt to acquire Ferrari, Henry Ford II resolved to challenge Ferrari in the prestigious 24 Hours of Le Mans race. To disrupt Ferrari's dominance, Ford's chief engineer, Roy Lunn, conceived the idea for the GT40. It was produced in limited runs from 1964 to 1969.
            \end{itemize}
            
        Case 2: Partial match. The meaning of $A'$ partially overlaps with the meaning of $A$, but not fully. For example, the following $A'$ generated using BM25 receives 3 points. Reason: $A'$ clearly leaves out information that the Ford GT40 was conceived by Ford's chief engineer, Roy Lunn.
            \begin{itemize}
                \item $Q$: What was the reason behind the creation of the Ford GT40?
                \item $A$: See Case 1 above.
                \item $A'$: The reason behind the creation of the Ford GT40 was to compete against Ferrari in racing events, as evidenced by Ford's continued efforts to improve the GT40 and best the Italians.
            \end{itemize}

    \item 0 points: $A'$ receive 0 points if $A$ and $A'$ are distinct from each other with no overlap in meaning. For example, the following $A'$ generated through EDI receives 0 points.
    \begin{itemize}
        \item $Q$: What was the reason behind the creation of the Ford GT40?  
        \item $A$: See Case 1 above. 
        \item $A'$: The Ford GT40 was created to take full advantage of the benefits associated with a mid-engine design, including a slinky aerodynamic shape and benign handling characteristics."              
    \end{itemize}
    \end{itemize}

Table \ref{table:extrinsic} shows the average scores of comparing answers by freelance annotators as well as the corresponding BERTSim scores. The integers in the row below the row of evaluation scores represent the respective rankings.

\begin{table}[h]
\caption{Evaluation scores by extrinsic evaluation and intrinsic BERTSim metric with rankings, where ``Extr" stands for ``extrinsic evaluation" and ``Intr" for ``intrisic evaluation"}
\label{table:extrinsic}
\includegraphics[width=\columnwidth]{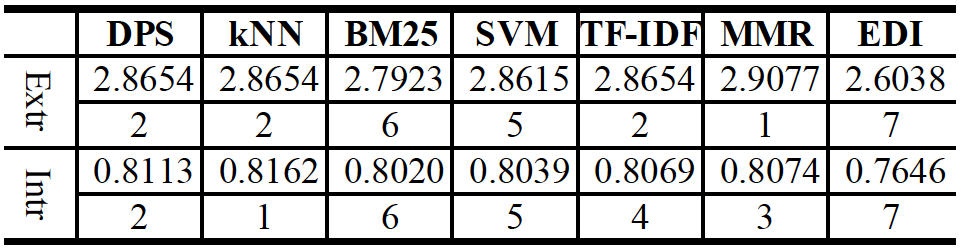}
\end{table}

It is evident that the extrinsic evaluation scores align well with the BERTSim scores, demonstrating consistency in ranking. In particular, both evaluations are in complete agreement for the 2nd, 5th, 6th, and 7th places, with only minor variations in the other rankings. For instance, MMR is ranked 1st by extrinsic evaluation and 3rd by BERTSim, which is quite close. Similarly, TF-IDF is ranked 2nd by extrinsic evaluation and 4th by BERTSim. Notably, DPS, kNN, and TF-IDF all share the 2nd rank in extrinsic evaluation, likely due to the coarseness of human annotation. Since DPS and kNN are essentially the same measures, they should logically be ranked closer to each other than to TF-IDF. Therefore, the extrinsic rank of TF-IDF, while differing slightly from BERTSim, can still be considered reasonably aligned.
Overall, this suggests a strong correlation between the two evaluation methods.

\subsection{Combining Multiple Retrievers}

LangChain supports combining multiple retrievers into a new retriever. We use the default setting to return four chunks for each combination. This approach diversifies the retrieved content from the reference document, potentially improving overall performance. 

As examples, we combine all seven retrievers, denoted as A-Seven; three retrievers with the highest OPI-1 scores: kNN, DPS, and TF-IDF, plus MMR for its strength in balancing relevance and diversity, denoted as A-Four; and two retrievers with the highest OPI-1 scores: kNN and DPS, denoted as A-Two. 

We may also combine the sentences retrieved by individual retrievers, removing any duplicates, and use the remaining set of sentences with the corresponding questions to generate answers and logical relations. Let S-Seven, S-Four, and S-Two denote the sets of sentences obtained this way by the corresponding retrievers as in A-Seven, A-Four, and A-Two.

The experimental results of both types of combinations are shown in
Table \ref{table:algcombination}.

\begin{table*}[]
    \centering
    \caption{Evaluation results of various combinations of retrievers and sentences}
    \includegraphics[width=\textwidth]{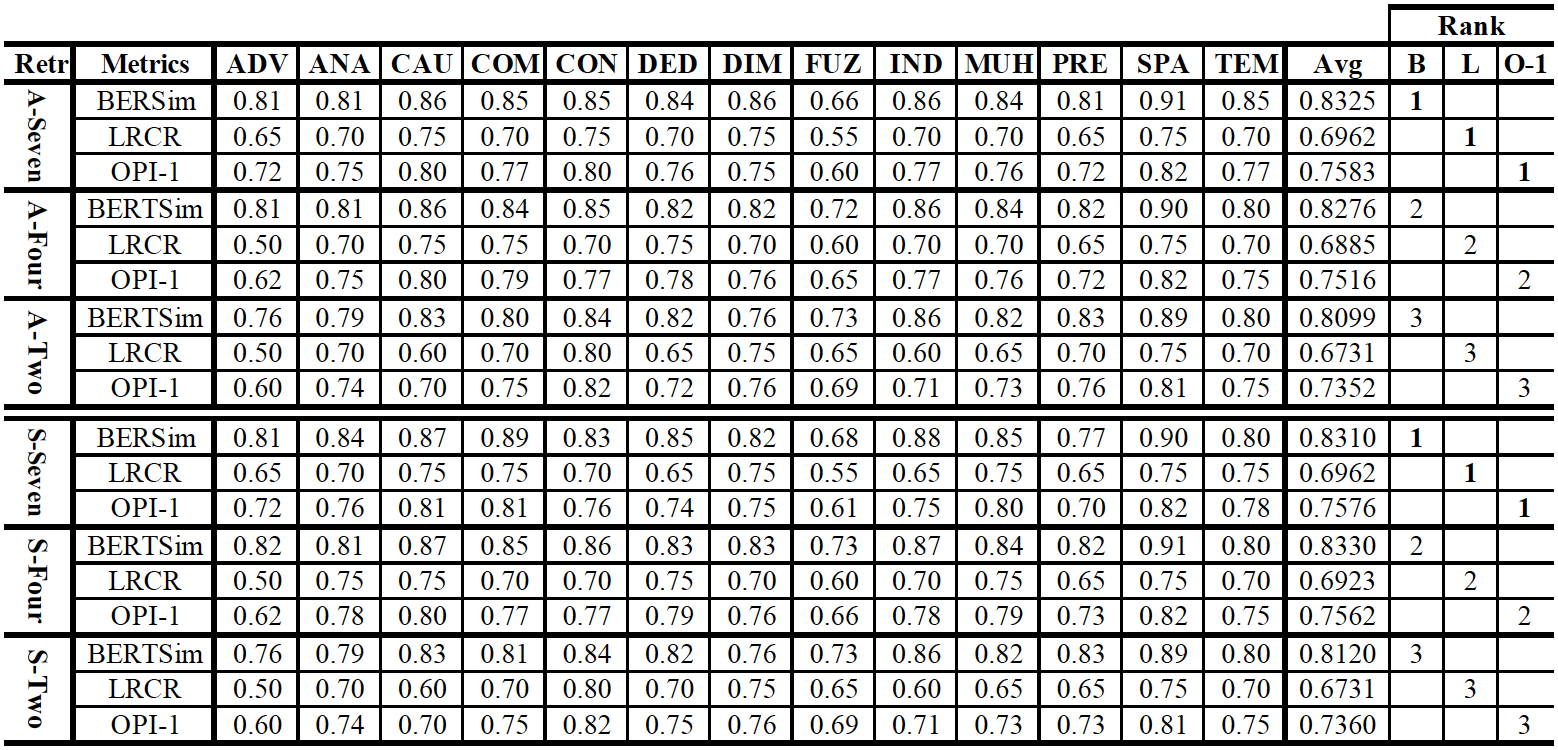}
    \label{table:algcombination}
\end{table*}

\section{\uppercase{Analysis}}
\label{sec:analysis}

We first analyze the performance of individual retrievers, followed by examining the combinations of retrievers and the sentences retrieved by multiple retrievers.

\subsection{Individual Retrievers}

For each retriever, we first analyze the performance for each logical relation individually and then assess the overall performance across all logical relations.

\subsubsection{Individual logical relation}
We use the OPI-1 scores to help identify the strengths and weaknesses of individual retrievers across the 13 logical relations. For example, as seen in Table \ref{table:intrinsic evaluation}, almost all retrievers tend to perform the worst on adversarial reasoning, followed by fuzzy reasoning. For other logical relations, the performance of retrievers varies, indicating that certain retrievers may be more suited to specific types of reasoning tasks while struggling with others. For example, 
even for the worst-performing retriever, EDI, which consistently ranks the lowest in both extrinsic and intrinsic evaluations of answer accuracy as seen in Table \ref{table:extrinsic}, it still performs best on deductive reasoning. This suggests that while EDI may generally be less effective across various logical relations, it has a particular strength in handling tasks that involve deductive reasoning. This example highlights the nuanced performance of retrievers, where even a generally weaker retriever can excel in specific logical tasks.
This variability in performance highlights the importance of selecting the appropriate retriever.

\subsubsection{Across all logical relations}

The average OPI-1 scores provide a means to identify, across all 13 logical relations, which retrievers are more suitable for specific tasks and which retrievers should be avoided. For example, as shown in Table \ref{table:intrinsic evaluation}, EDI has the lowest and SVM the second-lowest average OPI-1 scores, indicating they should generally be avoided. This is likely due to the limitations of the underlying features used to compute SVM scores and the coarseness of L2-norms when representing text chunks as bag-of-word vectors, which may fail to capture the nuanced relationships required for deep logical reasoning tasks. 

On the other hand, kNN has the highest and DPS the second-highest average OPI-1 scores, indicating that these retrievers would be the best choices for answering deep-logic questions. kNN (cosine similarity) and DPS are similar measures, with kNN being a normalized version of DPS, which explains their comparable performance. However, kNN takes slightly more time to compute than DPS, as DPS is the fastest among all seven retrievers---dot products are the simplest and quickest to compute compared to the operations used by other retrievers.

The MMR retriever allows GPT-4o to generate better answers across all logical relations, as shown in Table \ref{table:extrinsic}
}. However, it does not perform as well in producing the correct logical relations. This discrepancy may be attributed to MMR’s focus on balancing relevance and diversity in retrieved content, which improves answer quality but doesn't necessarily align with capturing accurate logical relations.

BM25 is in general more effective for retrieving longer documents in a document corpus with the default parameter values for $\kappa$ and $b$. However, to retrieve sentences from an article, it was shown that BM25 would should use different parameter values \cite{Zhang-Zhou-Wang2021}. 
This explains why BM25 is the second worse for generating answers as shown in Table \ref{table:extrinsic} by both extrinsic and intrinsic evaluations. It is not clear, however, why it produces a relatively higher LRCR value.

TF-IDF's performance falls in the middle range, which is expected. As a frequency-based approach, it may struggle to capture deeper semantic information, but it remains relatively effective because it retains lexical information, ensuring that important terms are still emphasized in the retrieval process.

\subsection{Performance of Various Combinations}

We first analyze the performance of combinations of retrievers versus individual retrievers, followed by an analysis of combining retrievers algorithmically versus combining sentences retrieved by individual retrievers within the combination.

\subsubsection{Combinations vs. individuals}

It can be seen from Table \ref{table:algcombination} that A-Seven outperforms A-Four, which in turn outperforms A-Two. A similar ranking is observed with S-Seven, S-Four, and S-Two. Moreover, both A-Seven and A-Four are substantially better than the top performer, kNN, when only a single retriever is used (see both Tables \ref{table:intrinsic evaluation} and \ref{table:algcombination}). A similar result is observed with S-Seven and S-Four, where combining more retrieved sentences from different retrievers also enhances performance, reinforcing the benefits of increased diversity in the retrieval process.
These results all confirm the early suggestion that combining more retrievers generally enhances performance in both algorithmic and sentence-based combinations, supporting the idea that diverse retrieval methods contribute positively to the overall effectiveness of the RAG system.

However, we also observe that some combinations of retrievers may actually lead to poorer performance compared to using the individual retrievers alone. This is evident in the case of A-Two and S-Two, the algorithmic and sentence combinations of kNN and DPS, both result in slightly lower average OPI-1 scores than kNN alone. This is probably due to the fact that kNN and DPS are very similar measures, and combining them doesn't significantly increase diversity. Worse, the extra information provided through their combination seems to have led to diminishing returns, negating the potential benefits of combining retrievers to improve performance. This phenomenon warrants further investigation.

Nevertheless, combining retrievers based on different retrieval methodologies could help increase diversity and, consequently, improve overall performance. This is evident in the case of A-Seven and S-Seven, which combine retrievers utilizing diverse retrieval methods, as well as in A-Four and S-Four, where MMR---a retrieval method that balances relevance and diversity---complements kNN. By leveraging varied retrieval techniques, we can ensure that a broader range of relevant content is retrieved, potentially leading to greater accuracy and more robust logical reasoning in the generated answers.

\subsubsection{Combining algorithms vs. combining sentences}

We compare the outcomes of combining retrievers at the algorithm level versus the sentence level. Combining retrievers at the algorithm level is a feature supported by LangChain, which returns the same default number of chunks before sentences are extracted. In contrast, combining retrievers at the sentence level involves merging sentences retrieved by individual retrievers, which may include more sentences than the algorithmic combination, and so should lead to a slightly better performance. This is evident when comparing A-Four with S-Four and A-Two with S-Two (see Table \ref{table:algcombination}). 

However, having more sentences may not always lead to improvement, as it can introduce conflicting information. This is evident when comparing A-Seven with S-Seven, where S-Seven has a lower average OPI-1 score than A-Seven. This is likely because A-Seven has already saturated the useful sentences, while S-Seven introduces additional sentences that negatively impact the average OPI-1 score.

In summary, these analyses suggest that, when combining appropriate retrievers, both algorithmic and sentence-level approaches offer performance improvements, with each method providing distinct advantages in terms of retrieval diversity and the quality of generated answers. Selecting appropriate retrievers requires a deeper understanding of the underlying retrieval mechanisms, making this an interesting topic for further investigation.

\section{\uppercase{Final Remarks}}
\label{sec:conclusion}

This paper presents an effective intrinsic evaluate method for the performance of RAG systems in connection to question-answering involving deep logical reasoning. 

LangChain supports a wide range of retrievers and allows users to integrate custom retrievers. Additionally, there are numerous large language models (LLMs) such as the Gemini series \cite{gemini2024}, LlaMA series \cite{llama2024}, and Claude series \cite{claude2024}, among others, as well as various retrieval-augmented generation (RAG) tools like LLAMAINDEX \cite{llamaindex2024}, HayStack \cite{haystack2024}, EmbedChain \cite{embedchain2024}, and RAGatouille \cite{ragatouille2024}.
Evaluating the performance of these models and tools, particularly for answering deep-logic questions where identifying logical relations is essential, represents an intriguing direction for future research.

Regularly reporting the findings of such investigations would significantly contribute to the advancement of RAG technologies. Furthermore, we aim to develop a tool that quantitatively assesses the depth of logical relations in question-answering systems relative to the underlying context. This effort would necessitate the creation of a new dataset that annotates the depth of each logical relation for every triple consisting of a question, an answer, and a set of reference sentences.

\bibliographystyle{apalike}
{\small
\bibliography{example}}



\end{document}